\documentclass[letterpaper]{article} 
\usepackage{aaai25}  
\usepackage{times}  
\usepackage{helvet}  
\usepackage{courier}  
\usepackage[hyphens]{url}  
\usepackage{graphicx} 
\usepackage{natbib}  
\usepackage{caption} 
\usepackage{algorithm}
\usepackage{algorithmic}

\usepackage{multirow}
\usepackage{amsmath}
\usepackage{booktabs}

\usepackage{newfloat}
\usepackage{listings}
\DeclareCaptionStyle{ruled}{labelfont=normalfont,labelsep=colon,strut=off} 
\floatstyle{ruled}
\newfloat{listing}{tb}{lst}{}
\floatname{listing}{Listing}
\title{Divide-and-Conquer: Tree-structured Strategy with Answer Distribution Estimator for Goal-Oriented Visual Dialogue}
\author{
    Shuo Cai\textsuperscript{\rm 1,2},
    Xinzhe Han\textsuperscript{\rm 3*},
    Shuhui Wang\textsuperscript{\rm 1,4}\thanks{Corresponding authors.}
}
\affiliations{
    \textsuperscript{\rm 1} Key Lab of Intell. Info. Process., Inst. of Comput. Tech., CAS, Beijing, China\\
    \textsuperscript{\rm 2} University of Chinese Academy of Sciences, Beijing, China\\
    \textsuperscript{\rm 3}China Academy of Aerospace Science and Innovation, Beijing, China\\
    \textsuperscript{\rm 4}Peng Cheng Laboratory, Shenzhen, China\\
    shuo.cai@vipl.ict.ac.cn, geraldhan@163.com, wangshuhui@ict.ac.cn
}

\usepackage{bibentry}

\begin{document}

\maketitle

\begin{abstract}
Goal-oriented visual dialogue involves multi-round interaction between artificial agents, which has been of remarkable attention due to its wide applications. Given a visual scene, this task occurs when a Questioner asks an action-oriented question and an Answerer responds with the intent of letting the Questioner know the correct action to take.
The quality of questions affects the accuracy and efficiency of the target search progress. However, existing methods lack a clear strategy to guide the generation of questions, resulting in the randomness in the search process and inconvergent results. We propose a Tree-Structured Strategy with Answer Distribution Estimator (TSADE) which guides the question generation by excluding half of the current candidate objects in each round. 
The above process is implemented by maximizing a binary reward inspired by the ``divide-and-conquer'' paradigm.
We further design a candidate-minimization reward which encourages the model to narrow down the scope of candidate objects toward the end of the dialogue. 
We experimentally demonstrate that our method can enable the agents to achieve high task-oriented accuracy with fewer repeating questions and rounds compared to traditional ergodic question generation approaches. Qualitative results further show that TSADE facilitates agents to generate higher-quality questions. 
\end{abstract}
\begin{links}
\link{Code}{https://github.com/caishuo-C/TSADE}
\end{links}

\section{Introduction}
Information-seeking through interaction is one of the most important abilities of artificial intelligence~\cite{matsumori2022study}. In recent years, goal-oriented visual dialogue has received increasing attention. This task is to generate questions by one agent and provide answer by users (or another agent) towards specific goal, as exemplified in many benchmarks, such as Guesswhat?!~\cite{de2017guesswhat}, GuessWhich~\cite{das2017learning} and VisDial~\cite{2017Visual}. 
The key of facilitating this task is how well the machine can mimic humans to raise questions about objects and seek answers about the visual scene. In fact, the efficiency and quality of the visual dialogue is measured by whether and how quickly the specific goal is achieved.

\begin{figure}[htbp]
  \centering
  \includegraphics[width=0.8\linewidth]{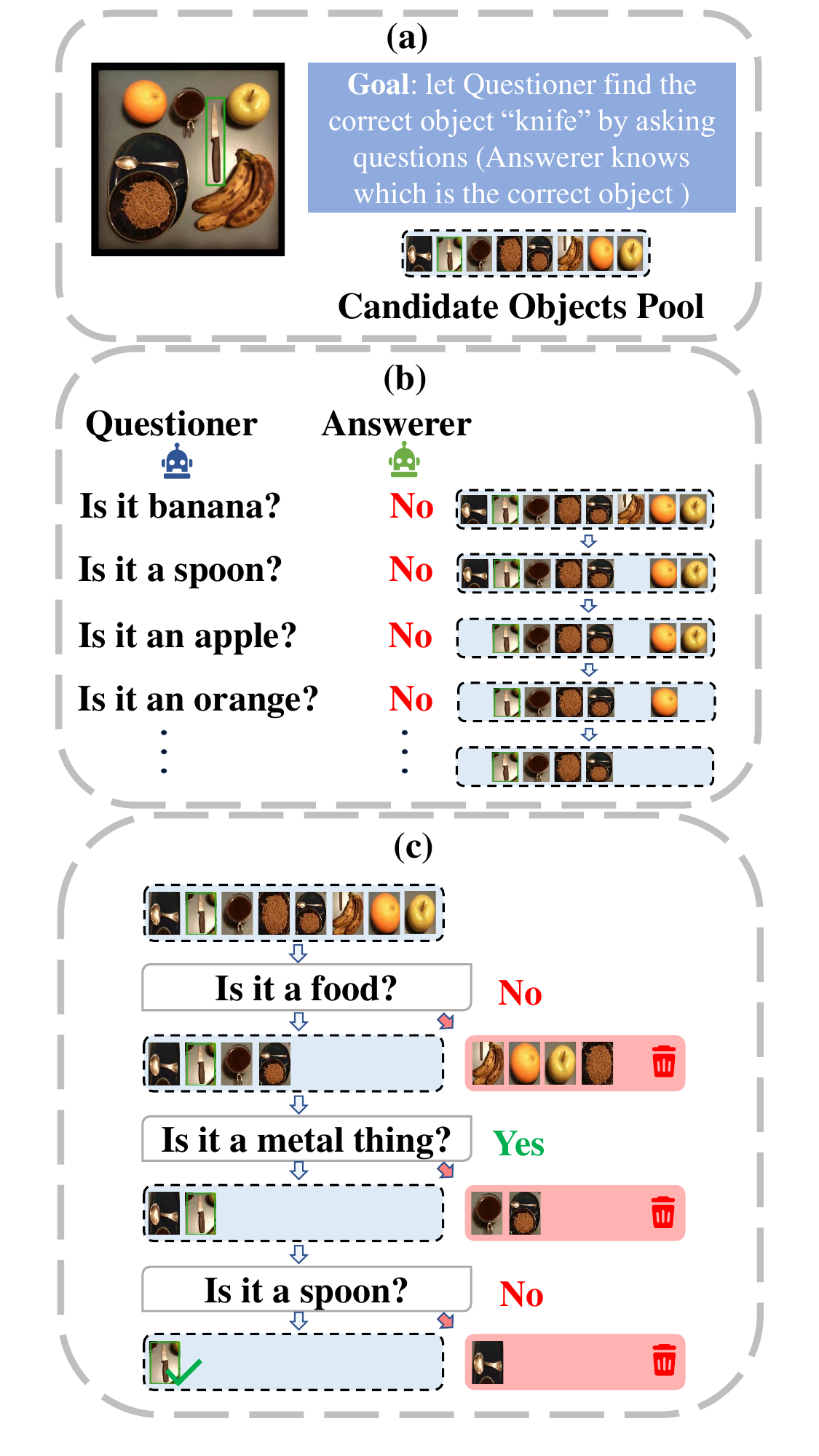}
  \caption{(a) Illustration of goal-oriented visual dialogue. The target object is highlighted in green box. (b) Example of traditional dialogue strategy. (c) Example of Tree-structure dialogue strategy. The excluded objects are in the lower-right candidate box.}
  \label{fig:1}
\end{figure}

See an example from GuessWhat?!~\cite{de2017guesswhat} in Figures \ref{fig:1} (a), the specific goal is to find the correct object (green box) from all objects (detected by object detection) in the image at the end. 
Towards this goal, the Questioner, responsible for generating question by its Question Generator (QGen) only seeing the whole image and the goal description, generates several rounds of questions, and Answerer, either machine Oracle in the training stage or human in testing stage, gives answer prompt by knowing what kind of objects beforehand. After multiple rounds of Q\&A, the Guesser, as another submodule of the Questioner, guesses which object is correct.
Obviously, the bottleneck problem in this process is the dialogue strategy, {\it i.e.}, how the questions are generated by QGen.


To enable agents to learn appropriate dialogue strategy and find the target object in a visual scene, mainstream solution is based on Reinforcement Learning (RL)~\cite{le2022multimodal} on benchmarks such as GuessWhat?!. 
These works~\cite{strub2017end,shukla2019should,zhang2018goal,abbasnejad2018active,abbasnejad2019s,zhao2018improving} usually design reward functions to achieve higher success rate or information gain based on the policy gradient~\cite{1999Policy}.
Other works improve the dialogue mechanism based on attention mechanism~\cite{anderson2018bottom} and the state of visual information in Supervised Learning (SL)~\cite{pang2020guessing, pang2020visual,tu2021learning}.
Despite the progress already made, existing mechanisms still need further investigation.
First, existing SL and RL methods only consider whether the target can be found in each round.
As shown in Figures \ref{fig:1} (b), their questioning strategies usually do not follow a clear path, thus questions are repeatedly and aimlessly asked to exclude a fraction of objects in each round. 
Second, even if the agent successfully identifies the target, the scope of the final candidate objects may not be reduced to a single object. 
The issue may be tolerable when there are only a few objects in the image. However, it would be difficult for the agent to find the target via aimless questions within limited rounds of Q\&A if there are dozens or even hundreds of objects, which is very common under real-world situations. In addition, the Questioner also encounters a high 
question repetition rate under these methods, lacking clear guidance for producing meaningful questions.

To address the above issues, in this paper, we propose a Tree-structured Strategy with Answer Distribution Estimator (TSADE) to guide Questioner to generate questions under RL paradigm.
In each round of Q\&A, Answer Distribution Estimator (ADE) employs a simulated Oracle that dynamically estimates the answer distribution of current candidate objects with the given question. Obviously, the objects that have the same answer as the ground truth (target object) have the potential to be selected as the target. With this answer distribution, we can constantly update the scope of candidate objects in the object pool.
We monitor candidate objects in order to calculate rewards under RL, which is independent of Guesser's final prediction. Even if the candidate objects are reduced to a a single object, the Guesser still predicts a target from all objects in the image.

We design two rewards to guide Questioner to learn TSADE for question generation, {\it i.e.}, \emph{binary} reward, and \emph{candidate-minimization} reward. The first reward is based on tree-structured strategy to perform ``divide and conquer''. Similar to the human decision-making process~\cite{stenning2012human}, in each round of Q\&A, we encourage the QGen to generate question that divides the current candidate objects into two groups with roughly the same number of objects. The grouping result should produce the greatest information gain compared to the previous round.  
Ideally, given $N$ candidate objects, this strategy can reduce the searching time complexity from $O(N)$ to $O(\log N)$.
We select the group where the ground truth (target) is located as the updated candidate objects in the next round. Although we use the ground truth mastered by Oracle, we do not break the information asymmetry between the Questioner and Oracle. We only use ground truth to calculate rewards when training QGen under the RL paradigm. In the inference stage, there is no need to get reward to calculate gradients to update model parameters, so the ground truth is not used. What needs to be emphasized is that when training QGen, we only use ground truth to maintain the correctness of candidate objects. The reward is calculated based on the changes in the overall distribution of answers. The ground truth information is not injected into the reward, and is not disclosed to the QGen that needs to be trained.
The second reward emphasizes the importance of dialogues that not only successfully finds the target but also narrows down the scope of candidate objects to only a single target at the end of the dialogue.
We encourage the models to find the target accurately, rather than finding it by a fluke.

We apply our method on various baselines on the GuessWhat?! and VisDial dataset. Experimental results show that our method can effectively improve the performance and reduce question repetition rate compared to numerous competitors on the datasets.

Our main contributions are as follows:
\begin{itemize}
\item We propose the Tree-structured Strategy with Answer Distribution Estimator (TSADE) to perform ``divide and conquer'' for goal-oriented visual dialogues. 
It helps the agent to find the target in fewer rounds.

\item TSADE can cooperate with various baselines to generate meaningful questions by reinforcement learning.

\item Experiments show that our method can help the agent to generate more informative questions and achieve the specific goal more quickly in visual dialogues.
\end{itemize}

\section{Related Work}
\subsection{Question Generator (QGen)}
The QGen plays a core role in the goal-oriented visual dialogue, as it not only needs to ask questions that can acquire certain information gain but also guides the dialogue towards the direction of the target.  
De Vries et al.~\shortcite{de2017guesswhat} propose the first QGen model with an encoder-decoder structure, in which the dialogue history is encoded by a Hierarchical Recurrent Encoder-Decoder (HRED)~\cite{serban2015hierarchical}, and the image is conditionally encoded as VGG features~\cite{simonyan2014very}.
Strub et al.~\shortcite{strub2017end} introduce the approach of RL and provide a 0-1 reward, where 1 indicates successful finding of the target in the dialogue. Built upon this approach, Zhang et al.~\shortcite{zhang2018goal} propose intermediate rewards from three dimensions to improve the model performance. 
Shekhar et al.~\shortcite{shekhar2018beyond} introduce a shared dialogue state encoder for Guesser and QGen, in which the visual encoder is based on ResNet~\cite{he2016deep}, and the language encoder is based on LSTM~\cite{hochreiter1997long}. Pang et al.~\shortcite{pang2020visual} introduce a turn-level object state tracking mechanism to QGen. Tu et al.~\shortcite{tu2021learning} introduce a Visual-Linguistic pre-trained model to QGen, which makes the object's semantic coverage more comprehensive and better.
Our main focus is on how to train QGen. 
The fundamental difference between TSADE and prior work lies in its clever use of a non-goal-oriented questioning strategy~(NGOQS) to find target, whereas prior works~\cite{zhang2018goal,shukla2019should,testoni2021looking} utilize a goal-oriented questioning strategy~(GOQS). 
We experimentally prove that flexibly using NGOQS is more useful than simply using GOQS, and GOQS can benefit from NGOQS.


\subsection{Answer Distribution Estimator (ADE)}
Given a question, ADE actually employs an internal Oracle to answer all objects in the image to obtain an answer distribution. Lee et al.~\shortcite{lee2018answerer} first introduce the ADE module to propose an Answerer in Questioner’s Mind (AQM) algorithm to obtain question in each round.
In this work, ADE refers to an approximated model of the original Oracle explicitly trained by AQM's Questioner. 
It abandons the paradigm of deep learning, and uses mathematics and the approximated model to directly calculate information gain to select question from training data in each round. 
Zhang et al.~\shortcite{zhang2018goal} propose three intermediate rewards to optimize the model in RL. 
Based on the goal-oriented way, it hope that the probability of ground truth (target) will progressively increase during the whole process. It uses ADE to avoid useless questions based on answer distribution. However, it does not consider what kind of questions are most useful. The difference is that TSADE takes the issue into account and uses ADE to achieve the same final goal in a non-goal-oriented way, without paying attention to which target is during the whole process.
Testoni and Bernardi \shortcite{testoni2021looking} propose the ``confirm-it'' strategy to select question that can gradually increase the probability of the target from the candidate questions. It uses an internal Oracle to provide answers specific to the target for a set of candidate questions. These answers are then used by the Guesser to compute a probability distribution over candidate objects. 


\section{Method}




\subsection{Background}



\textbf{Notations.} GuessWhat?! is a guessing game aiming to find the correct object from the image. Each instance of game is denoted as a tuple $\left( I, D, O, o^* \right) $, wherein $I$ represents the observed image, $D$ represents the dialogue consisting of $J$ rounds of Q\&A pairs $\left( q_j,a_j \right) _{j=1}^{J}$, and $O=\left( o_n \right) _{n=1}^{N}$ represents the list of $N$ objects in the image $I$, with $o^*$ referring to the target object. Each question $q_j=\left( w_{m}^{j} \right) _{m=1}^{M_j}$ is a sequence of $M_j$ tokens selected from a predetermined vocabulary $V$. The $V$ is comprised of word tokens, a question stop token $<\mathrm{?}>$, and a dialogue stop token $<\mathrm{End}>$. The answer $a_j\in \left\{ <\mathrm{Yes}>,<\mathrm{No}>,<\mathrm{NA}> \right\} $ can be categorized as either yes, no, or not applicable. 

\textbf{QGen.} The QGen produces a new question \(q_{j+1}\), given an image \(I\) and a history of \(j\) questions and answers \((q,a)_{1:j}\). It consists of a question encoder, an image encoder, and a question decoder. 

\textbf{Oracle.} The oracle is required to produce a yes-no answer $a_j$ for a target object $o^*$ within an image $I$ given a natural language question. The question is usually represented as the hidden state of an encoder, which is either LSTM or vision-language model~(VLM)~\cite{du2022survey}. 
We then concatenate the question, the bounding box of the target object, and category of the target object into a single vector and feed it as input to a single hidden layer multilayer perceptron~(MLP).
It outputs the final answer distribution using a softmax layer.

\textbf{Guesser.} The guesser takes an image \(I\), a history of questions and answers \((q,a)_{1:J}\), and predicts the correct object \(o^*\) from the set of all objects. The dialogue history is usually represented as the last hidden state of an encoder, which is either LSTM or VLM. The object embeddings are obtained from the categorical and spatial features. 
They are subjected to a dot product, which is then passed through a softmax to obtain a prediction distribution over the objects.

\textbf{SL Training.} The ground truths for QGen, Oracle, and Guesser are respectively the question, answer, and target object label. They are all optimized using cross-entropy loss.

\textbf{RL Training.} The Question Generation process is regarded as a Markov Decision Process (MDP), where the Questioner acts as an agent. At each time step $t$, for the generated dialogue based on the image $I$, the agent's state is defined as the visual information of the image, along with the history of Q\&A and the tokens of the current question generated so far: $S_t=\left( I,(q,a)_{1:j-1},\left( w_{1}^{j},...,w_{m}^{j} \right) \right) $, where $t=\sum_{k=1}^{j-1}{M_k}+m$. The agent's action $A_t$ involves selecting the next output token $w_{m+1}^{j}$ from $V$.
Depending on the agent's actions, the transition between two states falls into the completion of the current question, the end of the dialogue, or a new token.
The dialogue is limited to a maximum number of rounds $J_{\max }$. 
We model the QGen using a stochastic policy $\pi _{\theta}(A\mid S)$, where $\theta $ represents the parameters of the deep neural network utilized in the QGen baseline. These parameters are responsible for generating probability distributions for each state. The objective of policy learning is to estimate the parameter $\theta $.

\begin{figure*}[htbp]
  \centering
  \includegraphics[width=0.9\linewidth]{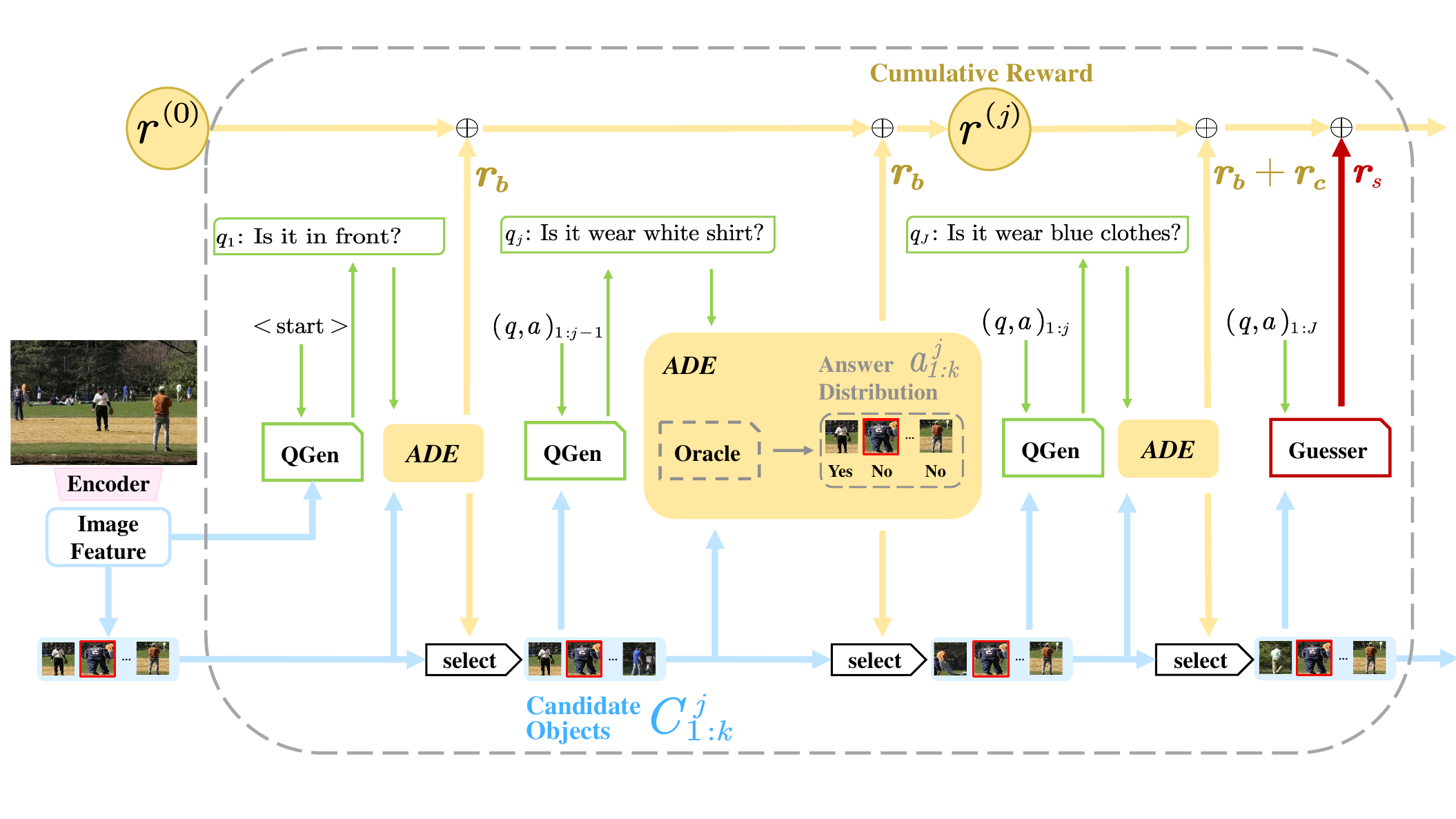}
  \caption{The framework of the Tree-structured Strategy with Answer Distribution Estimator (TSADE). The red box represents the target object.}
  \label{fig:2}
\end{figure*}

\subsection{Tree-structured Strategy with Answer Distribution Estimator (TSADE)}

\textbf{Tree-structured Strategy.} When humans are faced with this guessing game, they hope that each question can provide maximum distinction among the candidate objects. We propose a Tree-Structured strategy to mimic human behavior. As shown in Figure~\ref{fig:1} (c), while ensuring that no misclassification occurs, the Q\&A generated in each round should divide the current candidate objects into two groups as clearer as possible. In other words, half of the answers here should be ``Yes'' and the other half ``No''. The type of this question can be of any aspect, such as category, color, shape, size, location, and so on, as long as it meets the above requirement. Obviously, we also know which group the target belongs to. Therefore, we select this group as the new candidate object for the next round. 
Finally, after multiple rounds of questioning, we will find a single target. For example, in the second round, there are a total of four candidate objects, which can be divided into two groups, two metal things and two coffees. The Questioner asks the question ``Is it a metal thing?'' and the Oracle gives the answer ``No'', so two coffees are eliminated from the candidate objects.

\textbf{Answer Distribution Estimator (ADE).} As shown in Figure~\ref{fig:2}, ADE contains an Oracle to answer the question and obtains $a_{1:k}^{j}$ based on $C_{1:k}^{j}$ at each round. $C_{1:k}^{j}$ is a set of candidate objects maintained by ADE, which is updated in each round. $k$ refers to the total number of objects in the candidate objects. $a_{1:k}^{j}$ is an answer distribution aimed at $C_{1:k}^{j}$. $C_{1:k}^{j}$ is updated by selecting and only keeping the objects in $a_{1:k}^{j}$ that have the same answer as the target object. In addition, ADE outputs $r_b$ and $r_c$, which are \emph{binary} reward and \emph{candidate-minimization} reward based on the Tree-structured strategy. It measures $r_b$ and $r_c$ with $a_{1:k}^{j}$. It outputs $r_b$ in each round but outputs $r_c$ in the final round. When using the traditional RL paradigm~\cite{strub2017end}, a reward $r_s$ is given at the end. If the Guesser finds the target successfully, $r_s$ is 1, otherwise 0. $r_b$, $r_c$ and $r_s$ are added to the cumulative reward $r^{\left( j \right)}$ as the final reward.


\textbf{Binary Reward.} In a dialogue, to implement the Tree-structured strategy, we design the reward to measure whether we can eliminate half of the objects ($k/2$) to the maximum extent in each round of updating $C_{1:k}^{j}$. Given the state $S_t$, where the $<\mathrm{End}>$ token is sampled or the maximum round $J_{\max}$ is reached, the reward of the state-action pair is defined as follows:

\setlength{\abovedisplayskip}{3pt}
\begin{small}
\begin{equation}
r_b\left( S_t,A_t \right) =E\left[ \sum_{j=1}^{J_{end}}{\left( 1-\frac{|l_j-k_j/2|}{k_j/2} \right)} \right] 
\end{equation}
\end{small}
We score each round and calculate the overall expectation. $J_{end}$ refers to the round reaching $J_{\max}$ or the occurrence of $<\mathrm{End}>$ token. $k_j$ refers to the number of objects in $C_{1:k}^{j}$ in the $j$ round. $l_j$ has a value range of $\left[ 0,L_j \right] $, representing the maximum number of ``Yes'' or ``No'' in $a_{1:k}^{j}$. We hope that in each round of the dialogue, we can have $l_j=k_j/2$, where half of the answers are ``Yes'' and half are ``No''. In this case, the reward score for that round of dialogue is 1. When all answers are ``Yes'' or ``No'', the reward score is 0. 

\textbf{Candidate-minimization Reward.} Based on the process of finding the target in a dialogue, it is essentially equivalent to continuously narrowing down the scope of candidate objects. When the candidate objects are reduced to only one target, the goal is achieved. In many cases (especially when the dialogue has to stop due to reaching the maximum round limit $J_{\max}$), even if the target is successfully found in a dialogue, the scope of candidate objects has not been narrowed down to only the target. In this case, the result is not of high quality. In order to encourage the generation of higher-quality successful dialogues, we design a reward to measure the quality of those successful dialogues (compared with failed dialogues), as follows:

\setlength{\abovedisplayskip}{3pt}
\begin{small}
\begin{equation}
\begin{aligned}
r_c\left( S_t,A_t \right) =\begin{cases}
	\alpha 
 +\beta  
 \left( 1-\frac{k_{j_{end}}-1}{N-1} \right), &\mathrm{If\ } S_t \in \Omega, \\
	0, &\mathrm{Otherwise}.
\end{cases}
\end{aligned}
\end{equation}
\end{small}
where $\Omega=\{S|\arg\max_o[\mathrm{Guesser}(S)]=o^*\}$, $\alpha $ and $\beta $ are weights used to balance the contribution of the 0-1 reward and the quality of successful dialogues. When $k_{j_{end}}=1$, the dialogue is considered successful and of the highest quality. When $k_{j_{end}}=N$, we consider that even if the Guesser successfully finds the target $o^*$, the quality is not high, and it is likely a lucky guess. As $\alpha $ increases, the contribution of successful dialogues relative to failed dialogues to the overall reward increases. When the target $o^*$ is not successfully found, no reward is given.
When there is no 0-1 reward in this scenario, $r_c =\beta \left( 1-\frac{k_{j_{end}}-1}{N-1} \right) $. 
We encourage the generation of higher-quality successful dialogues by providing higher reward scores to those that can minimize the scope of candidate objects. 

\textbf{Training the QGen with Policy Gradient.} Given state-action pair $\left( S_t,A_t \right) $, we combine two rewards to form the ultimate reward function:

\setlength{\abovedisplayskip}{3pt}
\begin{small}
\begin{equation}
r\left( S_t,A_t \right) =\gamma \cdot r_b\left( S_t,A_t \right) +r_c\left( S_t,A_t \right) 
\end{equation}
\end{small}
where $\gamma$ is a weight to balance $r_b$ and $r_c$. Given the extensive range of actions in the game setting, we employ the policy gradient method~\cite{1999Policy} to train the QGen using the suggested rewards. This training method is similar to the approach used in the FS~\cite{strub2017end}.
\subsection{Applicability in Two Settings}

Our method can be treated as a plugin that can be applied to different models.
We divide the results into two settings for appropriate comparison.

\textbf{Without $r_s$.} If the models do not cooperate with TSADE, we only train the Oracle, Guesser, and QGen models independently using cross-entropy loss.
Then, if the models cooperate with TSADE, we keep the Oracle and Guesser models fixed and train the QGen model in the described RL framework with the rewards proposed by TSADE.
There is no 0-1 reward, which refers to $r_s$ from FS~\cite{strub2017end}.
FS can be considered as DV+$r_s$.
The purpose of the setting is to compare the model's results in SL with the model's results with TSADE in RL.
The training approach for QGen in SL follows the same details as DV~\cite{de2017guesswhat}.  

\textbf{With $r_s$.} We first independently train the Oracle, Guesser, and QGen models. Then keeping the Oracle and Guesser models fixed, we train the QGen model in the described RL framework using a comprehensive set of rewards. 
If the models follow the 0-1 reward proposed by FS, these rewards only include $r_s$.
On this basis, if the models cooperate with TSADE, these rewards consist of the $r_s$ and the rewards proposed by TSADE.
The purpose of the setting is to demonstrate the improvement results of TSADE on state-of-the-art models based on 0-1 reward in RL.


\section{Experiments}


\subsection{Dataset}

\textbf{GuessWhat?!.} We evaluate our method on the GuessWhat?! dataset~\cite{de2017guesswhat}, which contains 155k dialogs based on 66k images, including 134k distinct objects. The dataset consists of 821k question-answer pairs with a vocabulary size of 4900 words. We use the standard dataset split of 70\% for training, 15\% for validation, and 15\% for testing.

\textbf{VisDial.} To demonstrate the generalization of TSADE and its broader application scope, we additionally conduct experiments on the VisDial dataset~\cite{2017Visual}.
The VisDial dataset includes images, and each with a caption and 10-round Q\&A. We evaluate the performance in a Questioner-Answerer image-guessing setting (namely GuessWhich~\cite{das2017learning})  on the VisDial v1.0.
Then the Guesser predicts the unseen image. 
Lastly, candidate images in the validation split are ranked based on their similarity to the prediction, and the rank of the target image is computed.
In each round, the change of similary score can be regarded as the bianary Q\&A in GuessWhat. In our experiments, increase in similarity is considered a ``yes'', and a decrease is considered a ``no'' to implement our method.

\subsection{Evaluation Metrics}

\textbf{GuessWhat?!.} Following previous studies~\cite{de2017guesswhat}, we report the success rate of end-to-end models 
for different rounds in New Game or New Object. New Game refers to situations where the image and target have never been seen before. The New Object refers to an image that has been seen before, but the target object within that image has not been seen before. We measure the quality of dialogue using the question repetition rate. 

\textbf{VisDial.} The evaluation metrics include: 1) MRR: mean reciprocal rank of the target image; 2) R@k: the existence of target image in the top-k images.

\subsection{Implementation Details}

We conduct experiments on different QGen models (DV~\cite{de2017guesswhat}, FS~\cite{strub2017end} and ADVSE~\cite{2020Answer}).
We follow the source code of the above models.
ADVSE is a strong baseline that is comparable to the current state-of-the-art QGen. 
The QGen models are all combined with the Oracle-baseline~\cite{strub2017end} and Guesser-baseline~\cite{strub2017end} to complete end-to-end training. 
We follow all the experimental conditions of DV~\cite{de2017guesswhat} and FS~\cite{strub2017end}.
To make the comparison as fair as possible, we train all models for 40 epochs in SL or 150 epochs in RL with stochastic gradient descent (SGD)~\cite{bottou2010large}.
And we usually set the maximum number of rounds to 5.
The learning rate and batch size are 0.001 and 64. After the grid search, $\alpha$, $\beta$, $\gamma$ are set to 4, 0.7, 0.8.
When we train QGen, the image encoder is usually VGG~\cite{simonyan2014very}. So we replace the image encoder with CLIP~\cite{radford2021learning} and make a verification experiment.
And we make experiments on computational requirement analysis.

\begin{table}[!htbp]
  \centering
    \resizebox{1\linewidth}{!}{
    \begin{tabular}{l|ccc|ccc}
    \toprule
    \multicolumn{1}{c|}{\multirow{2}[4]{*}{\textbf{QGen (with $r_s$)}}} & \multicolumn{3}{c|}{\textbf{(\%)New Object}} & \multicolumn{3}{c}{\textbf{(\%)New Image}} \\
\cmidrule{2-7}          & \textbf{G} & \textbf{S} & \textbf{B} & \textbf{G} & \textbf{S} & \textbf{B} \\
    \midrule
    FS~\cite{strub2017end} & 66.27 & 51.13 & 67    & 66.64 & 51.2  & 66.8 \\
    ADVSE~\cite{2020Answer} & 42.19 & 38.06 & 41.41 & 42.21 & 38.63 & 41.5 \\
    \textbf{FS+TSADE} & \textbf{26.02} & \textbf{24.72} & \textbf{27.85} & \textbf{25.87} & \textbf{24.68} & \textbf{27.77} \\
    \textbf{ADVSE+TSADE} & \textbf{28.76} & \textbf{26.02} & \textbf{28.35} & \textbf{28.89} & \textbf{26.09} & \textbf{28.37} \\
    \bottomrule
    \end{tabular}%
    }
    \caption{Comparisons on repetition rate of the question.}
  \label{tab:2}%
\end{table}%

\begin{table}[ht]
  \centering
    
    \resizebox{1\linewidth}{!}{
    \begin{tabular}{c|l|c|cccc}
    \toprule
    \multirow{2}[4]{*}{\textbf{Settings}} & \multicolumn{1}{c|}{\multirow{2}[4]{*}{\textbf{QGen}}} & \multirow{2}[4]{*}{\textbf{Efficiency}} & \multicolumn{4}{c}{\textbf{number of rounds}} \\
\cmidrule{4-7}          &       &       & \textbf{5} & \textbf{6} & \textbf{7} & \textbf{8} \\
    \midrule
    \multirow{6}[4]{*}{without $r_s$} & \multirow{3}[2]{*}{DV} & T     & 3.35  & 3.56  & 3.74  & 3.88 \\
          &       & R(\%) & 26.46 & 28.79 & 30.4  & 31.43 \\
          &       & T/R   & 12.67 & 12.38 & 12.33 & 12.37 \\
\cmidrule{2-7}          & \multirow{3}[2]{*}{\textbf{DV+TSADE}} & T     & 3.62  & 3.86  & 4.06  & 4.22 \\
          &       & R(\%) & \textbf{30.79} & \textbf{34.16} & \textbf{36.49} & \textbf{38.07} \\
          &       & T/R   & \textbf{11.77} & \textbf{11.3} & \textbf{11.12} & \textbf{11.09} \\
    \midrule
    \multirow{6}[4]{*}{with $r_s$} & \multirow{3}[2]{*}{FS} & T     & 3.67  & 4.03  & 4.17  & 4.23 \\
          &       & R(\%) & 39.4  & 46.46 & 48.76 & 49.59 \\
          &       & T/R   & 9.33  & 8.67  & 8.55  & 8.54 \\
\cmidrule{2-7}          & \multirow{3}[2]{*}{\textbf{FS+TSADE}} & T     & 3.86  & 4.26  & 4.37  & 4.43 \\
          &       & R(\%) & \textbf{42.85} & \textbf{52.71} & \textbf{55.09} & \textbf{55.92} \\
          &       & T/R   & \textbf{9.01} & \textbf{8.08} & \textbf{7.95} & \textbf{7.92} \\
    \bottomrule
    \end{tabular}%
    }
    \caption{Comparison on efficiency.}
  \label{tab:table5}%
\end{table}%

\subsection{Comparison of Question Repetition Rate}

Although traditional methods have greatly improved the end-to-end accuracy with $r_s$ in the RL framework, a common issue is the high question repetition rate. 
As shown in Table~\ref{tab:2}, TSADE significantly reduces the question repetition rate for each QGen. 
G/S/B refer to the three methods of sampling questions: greedy, sampling, and beam-search (beam size 5). 
TSADE can improve the result of the FS by up to 40.77\% and the ADVSE by up to 13.43\%. 
These significant improvements clearly demonstrate the superiority of the proposed Tree-structured strategy. 
When question generation is not limited to goal-oriented, its scope is broader and diversity is better.





\subsection{Comparison on Efficiency}

As shown in Table~\ref{tab:table5}, T represents the average number of rounds to reach a single candidate object, while R represents the proportion of games where a single candidate object is reached.
We hope to reach a single candidate object in fewer rounds. When we measure the ability, smaller T and higher R indicate better results. 

However, the games that can reach a single candidate object under different models are different and their base numbers are also different.
Apparently, it is unfair to simply compare the size of T. Therefore, we employ a coefficient T/R to measure this ability. Obviously, a smaller T/R indicates that the model is more efficient.
As evident from the results, with the TSADE enhancement, there is a reduction in T/R compared to the existing methods. It means that TSADE makes the models to achieve the goal more efficiently.

\begin{table}[htbp]
  \centering
    
    \resizebox{1\linewidth}{!}{
    \begin{tabular}{l|l|ccc}
    \toprule
    \multicolumn{5}{c}{\textbf{(\%)New Image}} \\
    \midrule
         \multicolumn{1}{c|}{\textbf{Settings}} & \multicolumn{1}{c|}{\textbf{QGen}} & \textbf{G} & \textbf{S} & \textbf{B} \\
    \midrule
    \multirow{8}[4]{*}{without $r_s$} & DV~\cite{de2017guesswhat}    & 42.16 & 39.41 & 45.3 \\
          & GDSE~\cite{shekhar2018beyond}  & 47.8  & -     & - \\
          & VDST~\cite{pang2020visual}  & 45.94 & 42.92 & - \\
          & ADVSE~\cite{2020Answer} & 48.43 & 49.98 & 45.87 \\
          & CSQG~\cite{shi2021category}  & 49.9  & -     & 48.1 \\
          & Vilbert~\cite{tu2021learning} & 52.5  & -     & - \\
\cmidrule{2-5}          & \textbf{DV+TSADE} & 42.97 & 43.78 & 43.78 \\
          & \textbf{ADVSE+TSADE} & \textbf{54.22} & \textbf{51.26} & \textbf{56.56} \\
    \midrule
    \multirow{7}[4]{*}{with $r_s$} & FS~\cite{strub2017end}    & 56.86 & 55.95 & 57.26 \\
          & Bayesian~\cite{abbasnejad2019s} & 59.8  & 59    & 60.6 \\
          & VQG~\cite{zhang2018goal}   & 60.7  & 59.8  & 60.8 \\
          & VDST~\cite{pang2020visual} & 64.36 & 63.85 & 64.44 \\
          & ADVSE~\cite{2020Answer} & 65.31 & 64.06 & 65.45 \\
\cmidrule{2-5}          & \textbf{FS+TSADE} & 61.36 & 59.82 & 61.41 \\
          & \textbf{ADVSE+TSADE} & \textbf{65.99} & \textbf{64.45} & \textbf{65.94} \\
    \bottomrule
    \end{tabular}%
    }
    \caption{Comparison with previous QGen models on the task success rate.}
  \label{tab:4}%
\end{table}%

\begin{table*}[h]
  \centering
  
  \resizebox{1.0\linewidth}{!}{
    \begin{tabular}{ccccc}
    \toprule
    \textbf{Oracle} & \textbf{Guesser} & \textbf{QGen} & \textbf{Acc} & \textbf{Acc(+TSADE)} \\
    \midrule
    GPT-4-vision-preview~\cite{yang2023dawn} & GPT-4-vision-preview~\cite{yang2023dawn} & GPT-4-vision-preview~\cite{yang2023dawn} & 48.14 & \textbf{51.33} \\
    \bottomrule
    \end{tabular}%
    }
    \caption{Comparison result on LVLM-based method.}
  \label{tab: Comparison result on LVLM-based method}%
\end{table*}%

\subsection{Comparison on Success Rate of End-to-end}

We show the task success rate of different models with/without our method at maximum round 5. 
As shown in Table~\ref{tab:4}, compared with ADVSE model without $r_s$, the ADVSE+TSADE improves the task success rate by 1.72\% on greedy case in New Image. 
It indicates that TSADE can improve the task success rate of the model~(SL) after RL training.
Compared with the ADVSE with $r_s$, TSADE achieves a 0.68\% improvement and on greedy case in New Image.
In New Image, TSADE achieves a new state-of-the-art task success rate under RL.
It indicates that TSADE can also improve the task success rate of the model~(RL).


In addition, we use a large visual-language model (LVLM) to test the effectiveness of TSADE.
We use GPT-4-vision-preview~\cite{yang2023dawn} API as Oracle, Guesser, and QGen respectively, and use preset prompts to let three agents understand their roles and task settings. 
We prompt QGen with questioning strategy by TSADE to generate questions. This strategy involves:
\begin{itemize}
    \item a) In each round, the question is raised based on the concept of binary search, aiming to eliminate half of the remaining set of candidates in the image when answered.
    \item b) The goal is to narrow down the set of remaining candidates to as few objects as possible.
\end{itemize}
As shown in Table~\ref{tab: Comparison result on LVLM-based method}, TSADE still achieves constant improvement against the LVLM-based baseline. 
In addition, it is interesting to note that LVLM performs much worse than traditional small models on this task. Although LVLM has strong visual understanding ability, the goal-oriented visual dialogue task requires the model to have good spatial reasoning ability, which may be the ability that LVLM lacks~\cite{kamath2023s}.

\begin{table}[t]
  \centering
    
    \resizebox{0.85\linewidth}{!}{
    \begin{tabular}{l|cccc}
    \toprule
    \multicolumn{5}{c}{\textbf{(\%) Game Success Rate}} \\
    \midrule
    \multicolumn{1}{c|}{\multirow{2}[2]{*}{\textbf{Models}}} & \multicolumn{2}{c}{\textbf{New Object}} & \multicolumn{2}{c}{\textbf{New Image}} \\
          & \textbf{G} & \textbf{S} & \textbf{G} & \textbf{S} \\
    \midrule
    TSADE (full model) & 62.06 & 60.99 & 61.36 & 59.82 \\
    w/o $r_b$ & 60.09     & 58.28     & 59.52     & 57.69 \\
    w/o $r_c$ & 56.84     & 55.80     & 56.13     & 54.93 \\
    \midrule
    \multicolumn{5}{c}{\textbf{(\%) Question repetition Rate}} \\
    \midrule
    TSADE (full model) & 26.02 & 24.72 & 25.87 & 24.68 \\
    w/o $r_b$ & 39.93     & 39.80     & 39.86     & 39.26 \\
    w/o $r_c$ & 26.16     & 22.02     & 25.56     & 21.64 \\
    \bottomrule
    \end{tabular}%
    }
    \caption{Experimental results of ablation studies.}
  \label{tab:3}%
\end{table}%

\begin{table}[t]
  \centering
    
    \resizebox{1\linewidth}{!}{
    \begin{tabular}{c|ll|cccc}
    \toprule
    \textbf{Settings} & \multicolumn{2}{c|}{\textbf{Models}} & \textbf{MRR} & \textbf{R@1} & \textbf{R@5} & \textbf{R@10} \\
    \midrule
    \multirow{2}[2]{*}{without $r_o$} & \multicolumn{2}{l|}{ReeQ+AugG} & 31.21  & 17.78  & 45.01  & 59.98  \\
          & \multicolumn{2}{l|}{\textbf{ReeQ+AugG+TSADE}} & \textbf{33.21 } & \textbf{19.60 } & \textbf{47.04 } & \textbf{61.68 } \\
    \midrule
    \multirow{2}[2]{*}{with $r_o$} & \multicolumn{2}{l|}{ReeQ+AugG} & 33.65  & 19.91  & 48.50  & 62.94  \\
          & \multicolumn{2}{l|}{\textbf{ReeQ+AugG+TSADE}} & \textbf{34.30 } & \textbf{20.78 } & 48.11  & \textbf{63.61 } \\
    \bottomrule
    \end{tabular}%
    }
    \caption{Comparison results with/without TSADE based on VisDial v1.0.}
  \label{tab:re1}%
\end{table}%

\begin{figure}[htbp]
  \centering
  \includegraphics[width=1\linewidth]{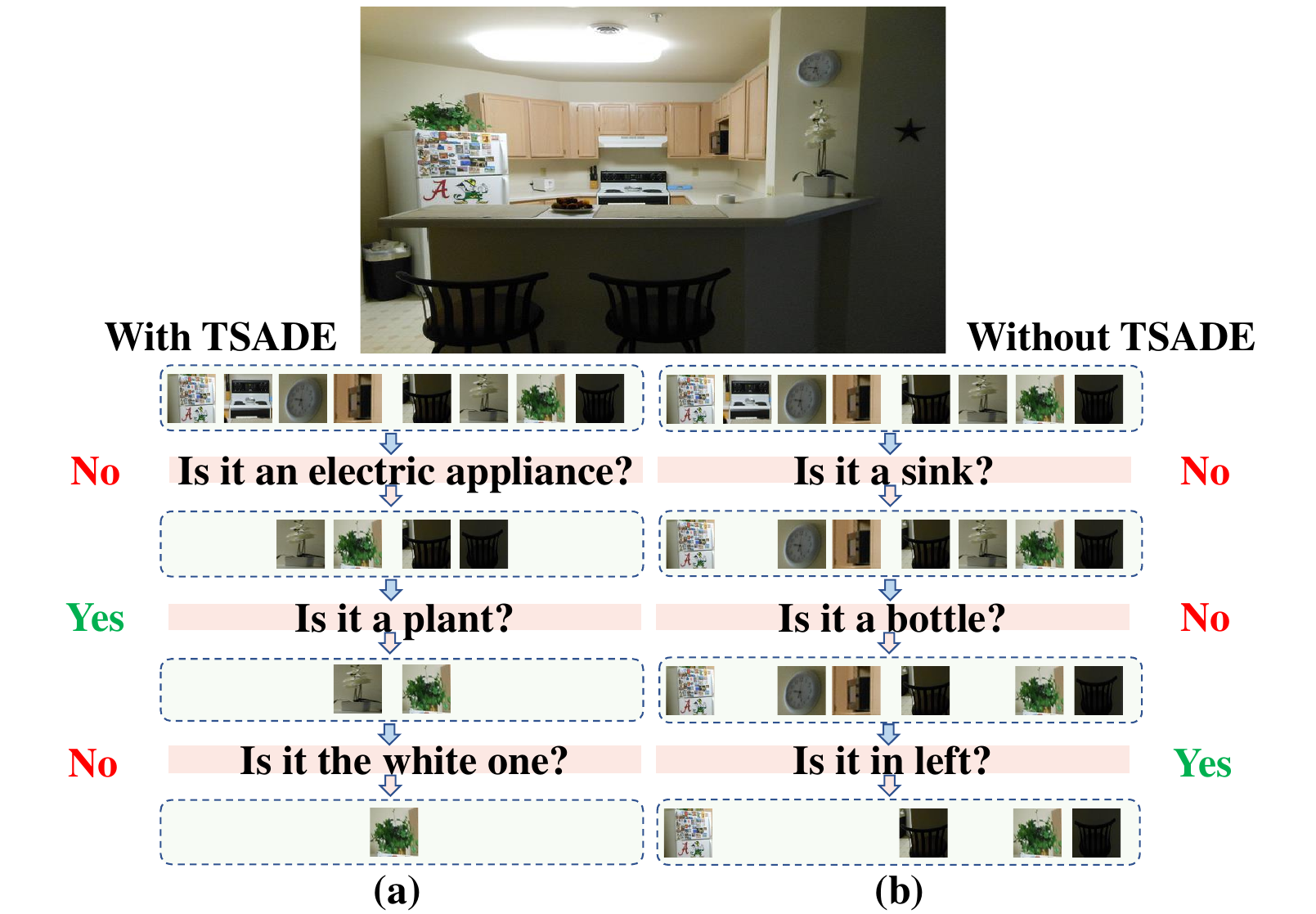}
  \caption{The generated dialogue examples show the strategy for question generation with and without TSADE.}
  \label{fig:5}
\end{figure}

\subsection{Ablation Study}

In Table~\ref{tab:3}, we evaluate the individual contribution of the 
\emph{binary} reward ($r_b$) and \emph{candidate-minimization} reward ($r_c$). We conduct ablation studies based on the FS~\cite{strub2017end} with $r_s$ at maximum round 5. 
In terms of the question repetition rate, the repetition increases by around 14\% without $r_b$, while it slightly decreases without $r_c$. Regarding the game success rate, it decreases by approximately 2\% without $r_b$ and 5\% without $r_c$. It indicates that $r_b$ not only improves the success rate to some extent but has a greater impact on reducing question repetition. On the other hand, $r_c$ primarily helps the model to significantly improve success rate by selecting higher-quality successful dialogues under RL. It can help the model identify the target object precisely.

\subsection{Qualitative Analysis of the Strategy}

In Figure~\ref{fig:5}, we present an example of a dialogue with/without TSADE for qualitative analysis. We only show examples with eight objects. As shown in Figure~\ref{fig:5} (a), when the model asks the first question, the feature represented by ``electric appliance'' matches four of the eight objects. After receiving the answer ``No'' from the Oracle, half of the objects are excluded. In the second round, the model selects two plants and two chairs from the candidate objects and asks a question about a shared feature of some objects, rather than asking about the specific feature of each object. After receiving the answer ``Yes'', two chairs are eliminated. Finally, the model asks a question about one of the candidate objects to identify the target object. In the Figure~\ref{fig:5} (b), the model tends to query ergodically, and a small number of objects can be eliminated in each round. It is very inefficient.

\subsection{Generalization and Limitation Analysis}


To demonstrate the generalization of TSADE, we provide experiments on VisDial v1.0~\cite{2017Visual} with the ReeQ~\cite{Zheng_Xu_Meng_Wang_Wang_Zhou_2021} model for GuessWhich~\cite{das2017learning} task.
$r_o$ refers to the original reward in ReeQ under RL.
As shown in Table~\ref{tab:re1}, it could be seen that our method achieves  performance improvements under most metrics.
It further verifies that our method could be generalized to other datasets. 
Furthermore, the thought also has potential exploration space in non-dialogue goal-oriented tasks. 
For example, we can try goal navigation~\cite{chaplot2020object}, where a set of candidate paths are established in the navigation process.
Based on whether the probability of each path increases or decreases, one can use ``yes" or ``no" to divide candidate paths set into halves. 

\section{Conclusion}

This paper presents a Tree-structured Strategy with Answer Distribution Estimator for goal-oriented visual dialogue. Following the ``divide and conquer'', TSADE generates question to exclude half of the candidate objects based on the answer distribution provided by the ADE in each round. We design two rewards within the RL framework to implement this strategy. The \emph{binary} reward aims to find the target with fewer round, while the \emph{candidate-minimization} focuses on selecting higher-quality successful dialogues. Experimental results on the GuessWhat?! and VisDial dataset show that our method can effectively generate more useful questions and improve the accuracy of the model. 
However, while existing visual-language models perform well at processing visual-language and abstract thinking, they are still incapable of understanding the physical world and performing practical planning.
In future work, we aim to study the goal-oriented visual dialogue towards embodied AI agents and applications, and pay special attention to the question generation techniques of machines to facilitate more comprehensive collaboration between human and AI agents.

\section{Acknowledgements}

This work was supported in part by the National Key R\&D
Program of China under Grant 2023YFC2508704, in part by
National Natural Science Foundation of China: 62236008,
U21B2038 and 61931008, and in part by the Fundamental
Research Funds for the Central Universities.

\bibliography{aaai25}

\end{document}